\documentclass[pmlr,twocolumn]{jmlr}

\usepackage[counterclockwise, figuresleft]{rotating} 
\usepackage{csquotes} 
   
\usepackage{longtable}

\usepackage{booktabs}
\usepackage[load-configurations=version-1]{siunitx} 

\theorembodyfont{\upshape}
\theoremheaderfont{\scshape}
\theorempostheader{:}
\theoremsep{\newline}

\jmlrvolume{193}
\firstpageno{1}

\jmlryear{2022}
\jmlrworkshop{Machine Learning for Health (ML4H) 2022}

\title[Meta-analysis of individualized treatment rules]{Meta-analysis of individualized treatment rules via sign-coherency}

  \author{\Name{Jay Jojo {Cheng}} \Email{jay.cheng@wisc.edu}\\
  \addr Department of Biostatistics and Medical Informatics, University of Wisconsin--Madison
  \AND
  \Name{Jared D. {Huling}} \Email{huling@umn.edu}\\
  \addr Division of Biostatistics, University of Minnesota
  \AND
  \Name{Guanhua Chen} \Email{gchen25@wisc.edu}\\
 \addr Department of Biostatistics and Medical Informatics, University of Wisconsin--Madison
  }

\begin{document}

\maketitle

\begin{abstract}
Medical treatments tailored to a patient’s baseline characteristics hold the potential of improving patient outcomes while reducing negative side effects. Learning individualized treatment rules (ITRs) often requires aggregation of multiple datasets(sites); however, current ITR methodology does not take between-site heterogeneity into account, which can hurt model generalizability when deploying back to each site. To address this problem, we develop a method for individual-level meta-analysis of ITRs, which jointly learns site-specific ITRs while borrowing information about feature sign-coherency via a scientifically-motivated directionality principle. We also develop an adaptive procedure for model tuning, using information criteria tailored to the ITR learning problem. We study the proposed methods through numerical experiments to understand their performance under different levels of between-site heterogeneity and apply the methodology to estimate ITRs in a large multi-center database of electronic health records. This work extends several popular methodologies for estimating ITRs (A-learning, weighted learning) to the multiple-sites setting.
\end{abstract}

\section{Introduction}
\label{sec:intro}
Two major trends in research and development across established pharmaceutical, biotech, and medical devices markets are the increasing use of real-world evidence (RWE) - observational data generated through routine clinical practice - and treatment benefit-risk analyses personalized to patients. Regulatory agencies in the U.S., Europe, and Japan have all issued guidance on the usage of RWE in supporting regulatory submissions and have moved to encourage the role of RWE throughout the entire drug development life-cycle, beyond the current use focused in the post-authorization phase \citep{nishioka2022evolving,european2020ema,fda2019submitting}.

Concurrently, widespread recognition that the best treatment on average for a group of trial participants may not be the best treatment for a particular individual has led to a literature on estimating treatment decision rules specific to individuals, called \textit{individualized treatment rules} (ITRs) or individualized policies. These decision functions map patient baseline covariates to treatment options \citep{murphy2003optimal,zhao2012estimating,kosorok2019precision}. Learning optimal ITRs relies on detecting subgroup heterogeneities in interactions between covariates and treatment, and thus requires large samples. However, the main limitation in greater RWE utilization is data availability \citep{flynn2020ability}, which can be even more pronounced in many therapeutic areas with small patient populations (e.g. rare disease, elderly, pregnant women). The need for large samples coupled with limited data availability leads to a core tension in ITR estimation: pooling datasets between sites is common practice, but existing ITR methodology assumes an idealized sample, wherein all observations are drawn independently from a single population. Furthermore, a single globally estimated treatment rule may not be flexible enough to reflect the optimal ITRs at each clinical site.

In light of these trends, we developed an individual-level meta-analysis (ILMA) framework for adapting existing ITR methodologies to the heterogeneous data setting. Our framework uses a scientifically motivated `directionality principle,' operationalized as an optimization penalty, to balance heterogeneity and information borrowing across sites, as well as a data-adaptive procedure for tuning site-specific models. Under our experiments, when sites are moderately heterogeneous, i.e. covariate shift and concept drift, our method outperforms both pooled approaches that estimate a single global model and separate learning approaches that use only site-specific information. Our analysis of transthoracic echocardiography use in a multi-center database also showcases its performance in sites with low data availability. Code for our implementation and scripts for reproducing all analyses are available at \url{https://github.com/jay-jojo-cheng/metacoop}.

\subsection{Related Work}
\label{sec:related_work}
ITR estimation is closely related to the problem of conditional average treatment effect estimation (CATE), for which there is a large and successful literature \citep{wager2018estimation,dorie2019automated,foster2019orthogonal,kennedy2020optimal,curth2021nonparametric,fan2022estimation}.
Indeed, with access to the true CATE, the Bayes optimal ITR is just the sign of the CATE. This motivates a class of ITR estimation methods, called \textit{indirect methods}, to first estimate the CATE and then invert these regression estimates for estimating ITRs. One potential drawback of this approach is that if the outcome models are misspecified, the estimators may not be consistent for estimating the optimal ITR within a given class. But ITR estimation and CATE estimation are different learning tasks, requiring different tradeoffs. For an extended perspective on the differences between CATE estimation and ITR estimation, see \citet{fernandez2022causal}.

Recognition of this has led to a literature on \textit{direct methods} for estimating ITRs. These methods are also called policy learning \citep{beygelzimer2009offset} or value-search \citep{zhang2015using} methods which target the clinical outcome and optimize for a decision rule with the best performance in a given class \citep{zhao2012estimating,zhu2017greedy}. Direct methods decouple the choice of models for the conditional outcome distribution from the choice of models for the ITR class so that the outcome distribution can be modeled flexibly while still controlling the complexity of treatment policy that will be implemented. Directly targeting treatment assignments over first estimating the CATE can also avoid misspecification issues while maximizing the performance of the ITR within the given class.

\citet{chen2017general} proposed a general statistical framework that unifies many direct methods in the binary treatment regime and our method can be seen as extending this framework to jointly learning different ITRs over multiple datasets. \citet{qi2020multi} proposed a method for ITR estimation in the multi-treatment regime and \citet{chen2016personalized} studied a continuous treatment (e.g. dose-finding).

The only other work to study ITR estimation under penalization that we are aware of is \citet{mbakop2021model} from the econometrics literature, which derives regret bounds of policy learning estimators with statistical penalties. However, differences with the current work include the zero-one loss function, single-dataset (IID) setting, and low-dimensional covariate in its application. The work also differs in spirit because the authors use penalization to achieve model selection, while our penalization is motivated by borrowing strength between studies.

We are not aware of any ITR methodology that addresses site heterogeneity arising from pooling multiple datasets. Within the literature on CATE estimation, \citet{tan2022tree} proposed a model-averaging method for improving CATE estimation in this regime. \citet{danieli2022preserving} proposed an indirect method for learning a single ITR from several sites that focuses on preserving data privacy at each site. Although the main goal of the current is borrowing strength between estimated ITRs to adapt to heterogeneity, in the appendix, we also show how our algorithms can be adapted to the setting of distributed datasets and compare it with existing literature on privacy-preserving meta-analyses.

\subsection{Notation and causal setup}
We denote the number of datasets (e.g. sites, hospitals, studies) as $K \in \mathbb{N}$. Within each of the $K$ studies, indexed by lowercase $k \in [K] = \{1, \ldots, K \}$,  we observe an IID sample of triplets $(Y^{k}_{i},X^{k}_{i},T^{k}_{i})$, drawn from some joint distribution $\mathbb{P}^{k}$. Here, $Y_{i}^{k}$ is the outcome variable (assume without loss of generality that smaller $Y$ is better), and $X_{i}^{k} \in \mathcal{X}^{k}$ is a $p$-dimensional random vector of covariate data, and $T_{i}^{k} \in \{-1, 1\}$ is control or treatment that is observed. We use $n^{k}$ to denote the total number of patients in the $k$th site and $i$ indexes the observation within the site. Distributions may differ between datasets, that is, $\mathbb{P}^{k} \neq \mathbb{P}^{k'}$. To be clear, we allow for heterogeneity in both $\mathbb{P}^{k}(x) \neq \mathbb{P}^{k'}(x)$ (covariate shift) and $\mathbb{P}^{k}(y^{(1)},y^{(-1)}|x) \neq \mathbb{P}^{k'}(y^{(1)},y^{(-1)}|x)$ (concept drift). When context allows, we may drop the superscript $k$ for notational brevity.

The $p$ features in $X_{i}^{k}$ are indexed by $j$ and indexing is such that the $j$th feature across the $K$ studies are similar. We will sometimes refer to the $j$th variables across all studies as the $j$th variable group. For simplicity, we only consider the case when each site has exactly one feature in the $j$th variable group, but extensions of our methodology to the general case are straightforward \citep{chiquet2012sparsity}. To summarize, we use indices (and total number) $i$ ($n^{ k }$) for observations, $j$ ($p$) for variables, and $k$ ($K$) for studies.

A treatment rule $d$ is a deterministic function over the space of covariates, mapping covariates to control or treatment: $d: \mathcal{X} \to \{-1,1\}$. We wish to estimate an individualized treatment rule $d^{k}$ for each site. We also adopt the potential outcome framework for causal inference \citep{rosenbaum1983central,rubin2005causal}. $Y^{k(1)}$ denotes the potential outcome related to receiving the treatment, and $Y^{k(-1)}$ denotes the potential outcome related to receiving the control. Let $\pi^{k}(t,x)=P(T^{k}=t|X^{k}=x) $ be the true propensity score for the $k$th site. We also adopt the following standard causal assumptions:
\begin{enumerate}
	\item Consistency/stable unit treatment value (SUTVA): $Y^{k}=I(T=1)Y^{k(1)}+I(T=-1)Y^{k(-1)}$
	\item Strong ignorability (no unmeasured confounders): $T^{k} \perp (Y^{k(1)},Y^{k(-1)})|X^{k}$
	\item Positivity: $0 < \pi^{k}(t,x) < 1$ for all $x$ that are observed.
\end{enumerate}

\section{ITR problem setup}
We briefly review the ITR learning problem in a single dataset setting via the \citet{chen2017general} framework before extending the formulation to the distributed learning over multiple datasets. We define the \textit{conditional average treatment effect} as $\Delta(x) = \mathbb{E}[Y^{(1)}-Y^{(-1)}|X=x] / 2$. This quantity reflects the average part of the outcome that is affected by treatment among patients with covariates $x$. We similarly define the \textit{conditional average main effect} as $\mu(x) = \mathbb{E}[Y^{(1)}+Y^{(-1)}|X=x] / 2$, which reflects the average part of the outcome that is not influenced by treatment. Algebraically, the conditional expected potential outcome can be expressed $\mathbb{E}[Y|T=t,X=x] = \mu(x) + t\Delta(x)$. We emphasize that this decomposition holds without any \textit{modeling} assumption on the relationship between features and outcome.

Our goal is to estimate a treatment rule, $d$, that minimizes the average loss in potential outcome (since lower is better) if used to assign treatment. We define the \textit{value function} $V(d)$ as $V(d) := \mathbb{E}[Y^{(d(X))}]$, which is the expected potential outcome from using decision rule $d$ to assign treatment. We define the optimal ITR as a decision function that minimizes the value function over a suitable class of decision rules, $\mathcal{D}$,
\begin{equation}
d^{\text{opt}} = \arg \min_{d \in \mathcal{D}} V(d). \label{optimalITR}
\end{equation}
We note that $-\text{sign}(\Delta(x))$ is the unconstrained minimizer of the value function over all possible decision rules (not just the ones in the class $\mathcal{D}$), but searching for this is combinatorially hard without any assumptions on the data distribution or form of $\Delta(x)$. Instead of using equation \eqref{optimalITR} to directly minimize for $d$, we identify $d^{\text{opt}}(\cdot) \in \{-1,1\}$ as the sign of a \textit{scoring function} $f^{\text{opt}}$ of the covariates $d^{\text{opt}}(x) = -\text{sign}(f^{\text{opt}}(x))$, with $f^{\text{opt}} : \mathcal{X} \to \mathbb{R}$. We can estimate $f^{\text{opt}}$ by using the following lemma due to \citet{chen2017general}:

\begin{lemma}\label{lem:loss}
Under the three causal assumptions above, $-\text{sign}(f^{\text{opt}})$ is an optimal individualized treatment rule if $f^{\text{opt}}$ is a minimizer of the population loss function,
\begin{equation}
    \mathcal{L}(f) = \mathbb{E}\left[ \frac{(Y - T f(X))^{2}}{\pi(T,X)} \right].
\end{equation}
\end{lemma}
We provide a full proof of this lemma in \appendixref{appendix:proof} for completeness. This lemma motivates the empirical two-step M-estimator
\begin{align}
    \arg \min_{f} \ & \mathcal{L}_{W}(f) \nonumber \\
    = \arg \min_{f} \ & \frac{1}{n} \sum_{i=1}^{n} \frac{(Y_{i} - T_{i}f(X_{i}))^{2} }{\hat{\pi}(T_{i},X_{i})}, \label{empiricWsq} 
\end{align}
where $\hat{\pi}(T_{i}, X_{i})$ is a consistent estimator of the propensity score, corresponding to \textit{weighted (covariate) learning} of \citet{tian2014simple}. Using similar arguments as in the proof of \lemmaref{lem:loss}, one can also conduct a similar derivation for \textit{A-learning} \citep{murphy2003optimal} and arrive at its corresponding loss function,
\begin{align}
    & \mathcal{L}_{A}(f) \nonumber \\
    = & \frac{1}{n} \sum_{i=1}^{n} \left(Y_{i} -  \{\frac{T_{i}+1}{2} - \hat{\pi}(1,X_{i}) \} f(X_{i}) \right)^{2}. \label{empiricAsq}
\end{align}

In the multiple site setting, our goal is to estimate separate ITRs $d^{1}, \ldots, d^{k}$ that minimize the \textit{multi-sites value function}, the sum of the expected potential outcomes under site-specific treatment rules:
\begin{equation}
    \sum_{k=1}^{K} V^{k }(d^{k }) = \sum_{k=1}^{K}\mathbb{E}[Y^{k(d^{k }(X^{k }))}]. \label{multistudyval}
\end{equation}
Recalling the identification of the optimal treatment rules as the negative sign of the conditional average treatment effect, the unconstrained minimizer of \eqref{multistudyval} can be identified as the minimizer of the \textit{multi-site loss function} (cf. Lemma \ref{lem:loss}), since  we can carry out the minimization for each summand separately: \begin{align}
    \sum_{k=1}^{K}\mathbb{E}\left[ \frac{(Y^{k} - T^{k} f^{k}(X^{k}))^{2}}{\pi^{k}(T^{k},X^{k})} \right]. \label{multistudyloss}
\end{align}

In the analysis of multiple sites, the conditional distribution of potential outcomes given baseline measurements (concept drift) and the distributions of $X$ (covariate shift) can be different; thus, the value functions can be different between sites. For example, consider the joint ILMA of two medical claims datasets, one with payment information from a private insurer and the other containing payment information from Medicare. Even though both outcomes are measured in dollars, they are best considered as different random variables. One Medicare dollar is not equal to one private payer dollar, with Medicare payments generally less than private insurer payments for the same procedure. Minimizing ITR loss functions over each of these datasets are related, but different, learning tasks, and the additive structure in equation \eqref{multistudyval} gives the flexibility to take this heterogeneity into account.

\subsection{Efficiency augmentation}
Minimization of equations \eqref{empiricWsq} or \eqref{empiricAsq} yield valid estimators of site-specific ITRs, but for a good choice of $\hat{a}(x)$, the augmented loss function 
\begin{align}
    \frac{1}{n} \sum_{i=1}^{n} \frac{(Y_{i}- \hat{a}(X_{i}) - T_{i}f(X_{i}))^{2} }{\hat{\pi}(T_{i},X_{i})},
\end{align}
can improve estimation efficiency in finite sample settings. The theoretically best augmentation function depends on $f^{\text{opt}}$, but because this optimal decision function is unknown and itself a target of estimation, in practice a working estimate \citep{tian2014simple} or a marginalization over the treatment is used \citep{chen2017general}. We pursue the latter here and improve estimation by using an `adequate' augmentation function that approximates $E[Y|X]$ over the samples with the form
\begin{align*}
    \hat{a}(X_{i}) = \sum_{t \in \{-1, 1 \}} \hat{\pi}(t,X_{i}) \hat{a}(t,X_{i}),
\end{align*}
where $\hat{a}(t,x)$ is an estimate of $E[Y | X=x, T=t]$. This choice is adequate in the sense that it is similar to the optimal augmentation function, thus it will generally lead to improvements in efficiency.
In the multiple sites setting, this augmentation function is estimated locally at each site, that is
\begin{align*}
    \hat{a}^{k}(X^{k}_{i}) = \sum_{t \in \{-1, 1 \}} \hat{\pi}^{k}(t,X^{k}_{i}) \hat{a}^{k}(t,X^{k}_{i}).
\end{align*}
We do so with random forests, but other regression approaches, e.g. splines, neural nets, can also be used for main effect augmentation in this framework.

\subsection{Comparison baselines}
Two natural comparison baselines to consider are locally learned models, where each site ignores data available at other sites and computes a `separate' model
\begin{align*}
    \hat{f}^{k}_{\text{separate}} = \arg \min L^{k}_{W}(f^{k}),
\end{align*}
where $L^{k}_{W}$ is the site-specific weighted learning loss function from \eqref{empiricWsq}. This gives each site the flexibility to learn its own ITR and should perform well when the optimal ITRs are quite different at each site. When each estimator is consistent, the joint estimator $\hat{f}_{\text{separate}} = (\hat{f}^{1}_{\text{separate}}, \ldots, \hat{f}^{K}_{\text{separate}})$ is also consistent for the multi-site loss function. The second comparison is a `pooled' model, where all samples are treated as a single dataset and a single ITR is jointly learned: \begin{align*}
    \hat{f}^{k}_{\text{pooled}} = \arg \min_{f} \sum_{k=1}^{K} \sum_{i \in [n^{k}]} \frac{(Y^{k}_{i} - T^{k}_{i}f(X^{k}_{i}))^{2} }{\hat{\pi}(T^{k}_{i},X^{k}_{i})}.
\end{align*}
The pooled model achieves optimal statistical error in the case where populations and decision functions are homogeneous across studies, but not when they can differ. Lastly, we note that local models may be computationally infeasible due to the high dimensional feature space and an insufficient number of samples in a single site.

\section{Directionality principle}
\subsection{Sign-coherent penalization}
To strike a balance between the flexibility of multiple treatment rules and the efficiency gains in pooling samples, we propose a directionality principle for borrowing information:

\begin{displayquote}
Locally, the same feature across studies should have the same directional contribution to its corresponding scoring function.
\end{displayquote}

As a concrete example, we have intuitively expect optimal site-specific decision rules to be smooth and directionally similar. We do not expect that small increases to blood pressure leads to a greater preference for the treatment in one location while small increases to blood pressure in another clinical location leads to a preference for the control.

We first formalize the above `directionality principle' in the special case of a linear scoring function, say $f^{k}(x) = \langle \beta^{k}, x \rangle$ for some parameter $\beta^{k}$, before we generalizing this to the case when $f$ comes from a weakly differentiable function class. We denote the concatenated parameter vector $\beta = \begin{bmatrix} \beta^{1\top} & \ldots & \beta^{K\top} \end{bmatrix}^{\top}$. The multi-site loss function for linear scoring functions becomes
\begin{align}
    L_{W}(\beta) = \sum_{k=1}^{K} \frac{1}{n^{k}} \sum_{i=1}^{n^{k}} \frac{(Y^{k}_{i}- T^{k}_{i} \langle \beta^{k} , X^{k}_{i} \rangle )^{2} }{\pi(T^{k}_{i},X^{k}_{i})} 
\end{align}
for weighted learning and 
\begin{align}
    L_{A}(\beta) = \sum_{k=1}^{K} \frac{1}{n^{k}} \sum_{i=1}^{n^{k}} (Y_{i}^{k} - \{ \frac{T_{i}^{k} + 1}{2} \nonumber \\
    - \pi(X_{i}^{k}) \}\langle \beta^{k} , X^{k}_{i} \rangle )^{2}
\end{align}
for A-learning. We add to the multi-site loss functions the following directionality penalization term:
\begin{align}
	\lambda P(\beta) = \lambda \sum_{j=1}^{J} (||(\beta_{j})_{+}|| + ||(\beta_{j})_{-}||  ),
\end{align}
where the norm is Euclidean and $(v)_{+}$ and $(v)_{-}$ return the componentwise positive and negative parts of the vector matching the particular sign. For example, if $\beta_{1}=\begin{bmatrix}1,2,3,-1,-2,-3 \end{bmatrix}$, then $(\beta_{1})_{+}=\begin{bmatrix}1,2,3,0,0,0 \end{bmatrix}$. The objective for weighted learning becomes
\begin{align}
    L_{W}(\beta) + \lambda P(\beta) \label{empiricWsqwpen}
\end{align}
and for A-learning becomes
\begin{align}
    L_{A}(\beta) + \lambda P(\beta). \label{empiricAsqwpen}
\end{align}
Here, we use $\beta_{j}$ to denote the length $K$ vector of the $j$th element from each site $\beta_{j} = \begin{bmatrix} \beta^{1}_{j} & \ldots & \beta^{K}_{j} \end{bmatrix}^{\top}$, which we call the $j$th coordinate block. In the specialization of our directionality principle to linear scoring functions, the optimization objective and penalty, $L(\beta) + \lambda P(\beta)$, can be interpreted as solving a weighted least squares objective with the cooperative LASSO penalty \citep{chiquet2012sparsity,huling2021two}. The geometry of the cooperative LASSO constraint encourages the $j$th coefficient of each site to match in sign such that they are zero and positive or zero and negative. This constraint encourages the same group of features to have an effect in the same direction and it penalizes large effects with opposing signs among similar features.

\subsection{Directionality principle for general function classes}
\label{directionality}
In the general case where $f$ comes from arbitrary function classes, we formulate the directionality principle with function norms of gradients. Here, the penalty added to the loss function takes the form:
\begin{align*}
    \lambda ||f||_{\mathbb{H}} \equiv & \lambda \sum_{j=1}^{p} \Bigg[ || (\begin{bmatrix}
        \partial_{j} f^{\{ 1 \}} \\
        \vdots \\
        \partial_{j} f^{\{ K \}} 
    \end{bmatrix})_{+} ||_{L_{2}}
    \\
    &
    + || (\begin{bmatrix}
        \partial_{j} f^{\{ 1 \}} \\
        \vdots \\
        \partial_{j} f^{\{ K \}} 
    \end{bmatrix})_{-} ||_{L_{2}} \Bigg],
\end{align*}
where $|| g ||_{L_{2}}$ is the $L_{2}$-norm $(E[g(X)^{\top}g(X)])^{1/2}$, $\lambda$ is a tuning parameter, and  $(v)_{+}$ and $(v)_{-}$ denote positive and negative parts of the vector $v$. Here we integrate the positive and negative parts of the gradient. When $f$ is a linear model, we recover the special case of the previous section. It is straightforward to check that this is indeed a norm and as suggested by the notation, it can be identified as the Hilbert norm of a reproducing kernel Hilbert space since it is bounded by two times the usual $L^{2}$ norm. Our learning problem with this generalized directionality penalty can be interpreted as a particular instance of kernel ridge regression, leading to statistical properties that are well understood \citep{vovk2013kernel, steinwart2009optimal}.

\subsection{Adaptive Information Criteria \label{CIC}}
The optimal choice of tuning parameter $\lambda$ reflects the degree to which sign-incoherence is penalized. Typically, hyperparameter tuning requires computationally intensive model refitting, as in cross-validation. Here, we use an adaptive variant of the concordance information criterion (CIC) and value information criterion approaches (VIC) \citep{shi2021concordance} which have the advantages of reusing computation from the augmentation functions and weights (thus skipping computationally intensive cross-validation) and directly targeting the ITR value function, as opposed to optimizing prediction error. We provide more details in \appendixref{appendix:adaptive}.

\section{Evaluation}
\subsection{Simulation study}
In evaluating the proposed methodology, we consider three main scenarios with low, medium, and high site heterogeneity. The basic scenario has moderate inter-site treatment effect heterogeneity in order not to favor either the pooled or separate learning approach. Scenarios two and three have high and low levels of inter-site treatment effect heterogeneity, respectively, where the separate analysis or a pooled analysis are alternatively favored. Each scenario is simulated in 18 sub-scenarios, for a total of 54 sub-scenarios, varying in the number of samples per site ($n=50, 100, 200$), the number of total predictors ($p=100,500$), and the number of studies ($K=3,5,10$). We pay special attention to these ``low'' data per site settings to mimic the situation in many aggregated databases: small amounts of site-specific data, but large in the aggregate. Moreover, this is the ``difficult'' regime for ITR methods. As will be seen in the results, sensible estimators generally perform well with enough data. Each sub-scenario is replicated 500 times. For space considerations, we show one sub-scenario for each of the scenarios in the main text and show the full figures in the appendix.

Model parameters for the data generation process are randomly drawn from heavy-tailed distributions and our goal is to see how the meta-analysis approach balances the homogeneity and heterogeneity between sites to either borrow strength or increase flexibility. We compare the methodology on relative value function, or the fraction of the maximum potential value captured by the treatment rule, and the accuracy of the treatment rule compared to the true sign of the treatment effect.

The estimation methods compared are two meta-analysis methods, with weighted and A-learning as the base learners; three centrally pooled methods, pooled weighted learning for a single model, pooled A-learning for a single model, A-learning with group lasso penalty to fit site-specific models; and two local methods, separate weighted learning, separate A-learning.

The first two meta-analysis methods use our overall method and a CIC step. The pooled weighted and A-learning analyses represent an idealized analysis with individual-level data sharing but a single global model is learned. We combine the $K$ datasets into a single dataset and estimate a single ITR using weighted or A-learning. A-learning with a group lasso penalty fits site-specific models constrained with a group lasso penalty, representing a pooled approach that attempts to incorporate site-specific tailoring under an alternative penalization. The separate analyses estimate $K$ ITRs on the datasets separately and represent local analyses done without any information sharing between models. The meta-analyses refer to the method described in this article.

Propensity scores and main effect augmentations are both fit with random forests using 5-fold cross-fitting (\appendixref{appendix:crossfit}). The same weights and augmentations are used across all methods in order to compare the treatment rule estimation scheme rather than nuisance parameter estimation. A detailed description of the data generation process is given in \appendixref{appendix:simulation}.

\begin{figure}[htbp]
\floatconts
  {fig:low_high_het}
  {\caption{Meta-analysis adapts to the level of site heterogeneity compared to base learners.}}
  {%
    \subfigure[Low site heterogeneity]{\label{fig:low_het}%
      \includegraphics[width=0.95\linewidth]{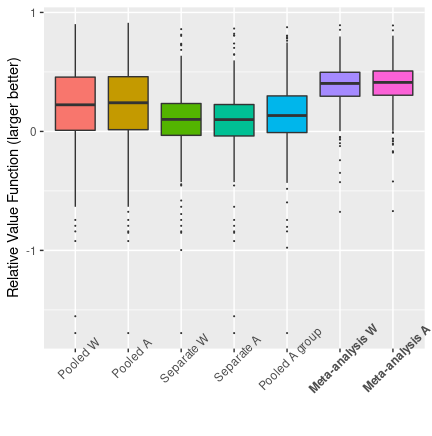}}%
    \qquad
    \subfigure[High site heterogeneity]{\label{fig:high_het}%
      \includegraphics[width=0.95\linewidth]{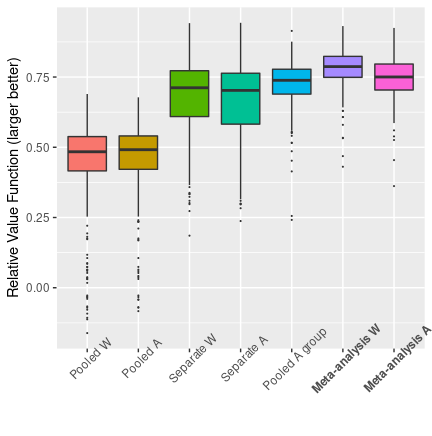}}
  }
\end{figure}

\subsubsection{Experimental findings}
Under low site heterogeneity, pooled methods are expected to outperform separate methods, because they are able to leverage a larger sample size. The sub-scenario shown in the main text typifies the general trends in the full set of simulations. Here we see that the meta-analysis methods perform comparably or even better. In particular, the variance seems to be smaller. In the full results of Figure \ref{fig:50val}, we can see that sometimes the performance of meta-analysis is able to quickly catch up as the amount of data increases. Both vastly outperform local separate learning analyses.

Figure \ref{fig:00val} shows a setting where the true treatment rules are moderately heterogeneous between sites. The blue dotted line indicates the average performance of the pooled oracle predictor within the rule class - fitting a site-specific treatment rule with access to infinite samples at each site. In a moderately heterogeneous setting, the centralized estimator that fits a single model still outperforms the site-specific local estimators. As the number of sites or dimensionality increases, the advantage of the meta-analysis over the local models improves.

Under high site heterogeneity, the pooled estimator underperforms against the local models. Separate learning methods are expected to outperform the pooled methods, because the optimal ITR at each site is quite different. Here the increase in sample size does not provide an advantage for estimation compared to the heterogeneity of the datasets. Here, we also see that meta-analysis is flexible enough to fit site-specific ITRs, while still borrowing strength to improve estimation, even though the sites are quite different. The full results in Figure \ref{fig:05val} show that although the true treatment rules are very different between studies, the meta-analysis method is still able to outperform the local models by taking advantage of sign-coherency.

The accuracy results (reported in the appendix) are similar to the relative value function but more pronounced. In terms of treatment accuracy, local approaches lag behind until the sample size increases. The accuracy of pooled methods also is extremely variable in low-data settings. This suggests that the pooled methods are more accurate on patients with larger conditional average treatment effects, but accuracy may be unstable in others.

\subsection{Individual-level meta-analysis of transthoracic echocardiography}
We apply the proposed methods above to reanalyze the ECHO study \citep{feng2018transthoracic} on the use of transthoracic echocardiography (TTE) in ICU patients with sepsis. Echocardiography can be used in critically ill patients to detect unsuspected cardiac abnormalities, but it is unclear whether or not care management arising from the echocardiography result alters patient outcomes. Feng et al. \citep{feng2018transthoracic} studied whether or not TTE use improves outcomes for patients on average, using the MIMIC III database of ICU patients from the Beth Israel Deaconess Medical Center in Boston, Massachusetts \citep{johnson2016mimic, goldberger2000physiobank}. Their analytic sample consists of 6361 patients and 40 baseline variables, including demographics, admissions data, vitals, labs, and other medical interventions, and their goal was to estimate the average causal treatment effect of using TTE. 

In our analysis, we replicate the ECHO study cohort within the publicly available eICU database \citep{goldberger2000physiobank}. Replicating the ECHO study with the eICU database typifies standard meta-analyses because eICU consists of 208 hospitals categorized into Midwest, Northeast, South, and West geographical regions. We thus split the data into four `sites' based on these regions. Noting that the MIMIC database does not contribute any patients toward eICU, our sample consists of 21,317 new patients and 38 baseline variables chosen based on the original study. We use our methodology to estimate a decision rule that determines whether or not TTE should be used on a given patient to minimize hospital length of stay. To account for outcome censoring in 11.4\% of patients, we use random survival forests \citep{ishwaran2008random} to impute discharge times with a procedure described in detail in \appendixref{supp:censimpute}. Less than 5\% of data cells in the baseline features were missing, and these were imputed using random forests \citep{tang2017random}. We reemphasize that our estimation target is no longer an average causal treatment effect as in the original study (causal effect estimation), but an \emph{individualized} treatment rule (causal decision-making).

\begin{figure}[htbp]
\floatconts
  {fig:echo_line}
  {\caption{Estimated ECHO study discharge times from using estimated treatment rules (lower is better).}}
  {\includegraphics[width = \linewidth]{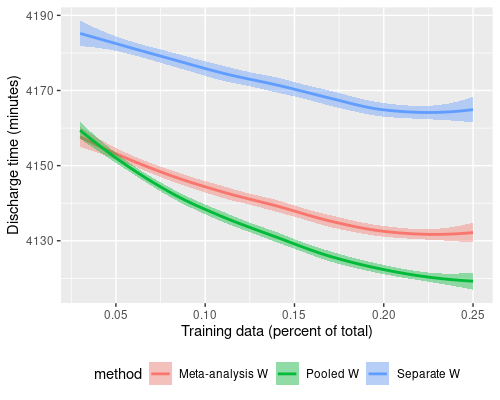}}
\end{figure}

We vary the amount of training data used from 2\% to 25\% and let the test set be the remainder of the data. This was done to assess performance in small data settings that commonly appear in biomedical applications. At each level of training data used, for 500 replications, the data is randomly split into training and test data. On the training sets, we run the proposed methodological pipeline, using energy balancing weights, weighted learning, and adaptive VIC to learn a meta-analysis model. We compare with local analysis, separate weighted learning, and a pooled weighted learning method with access to training samples from all sites. We compare the methodologies on the AIPW estimate of their value functions over the test data. The results are presented in \figureref{fig:echo_line}. The graph also suggests that our methodology considerably outperforms a locally learned model and results in improved discharge times of about 20-25 minutes compared to local methods and performs comparably to a method that pools all the data.

\section{Discussion}
In this article, we extended the ITR learning problem to the meta-analysis setting. Our proposed method leverages a directionality principle to borrow strength across multiple sites for learning ITRs by encouraging sign-coherency among site-specific ITRs and maintains flexibility by allowing them to be different. 

In simulation studies, we compared how our method performs against local site-specific models that do not leverage global information and global models with have access to all data but do not tailor decision rules to sites. Even in scenarios where settings favor these idealized models, our method performs comparably or even better. We further demonstrated the usefulness of our method by creating individualized transthoracic echocardiography rules on ICU discharge data from a network of hospitals.

The methodology has several natural extensions worth further investigation. The generalized penalization strategy given in the appendix can be applied to any learning algorithm that relies on optimizing a loss function (e.g. neural nets, support vector machines, etc.). One can also ``mix-and-match'' estimators within this more general formulation, say if separate sites model their data differently. The general formulation also has an equivalent expression as kernel ridge regression and this can be a convenient way to learn a more flexible treatment rule.

One weakness of the method is that sign-coherency is a difficult property to verify in practice. However, if sign-coherency does not hold, it may be an indication that these separate datasets should not be analyzed jointly to begin with, and one can fall back to the local estimator. Regardless, our approach seems to strike a balance when facing this uncertainty. In the future, we would also be interested in extending the work to the online learning regime.

\acks{JJC was supported by an NLM training grant to the Computation and Informatics in Biology and Medicine Training Program (NLM 5T15LM007359). Research reported in this work was partially funded through a Patient-Centered Outcomes Research Institute (PCORI) Award (ME-2018C2-13180). The views in this work are solely the responsibility of the authors and do not necessarily represent the views of the Patient-Centered Outcomes Research Institute (PCORI), its Board of Governors or Methodology Committee.}

\bibliography{cheng22}


\appendix
\section{Proof of \lemmaref{lem:loss}}
\label{appendix:proof}
A similar argument is sketched out in \citet{chen2017general}. We give an explicit proof here for completeness. In addition to the three causal assumptions, we will need to assume squared integrability of $Y$, that is $E[Y^2] < \infty$.
\begin{proof}
Let $L(\phi,\chi) = \mathbb{E}\left[ \frac{(Y - T \phi)^{2}}{\pi(T,X)} \big| X = \chi \right]$ be a function of two arguments from $\mathbb{R}^{2} \to \mathbb{R}$. This is well defined by positivity and squared integrability of $Y$. By the variational Euler-Lagrange equation, $f^{\text{opt}}(x)$ is a stationary trajectory of $\mathbb{E}_{X}[L(\phi,X)]$ if and only if $\frac{\partial L(\phi,\chi)}{\partial \phi}$ evaluated at $\phi = f^{\text{opt}}(x)$ and $\chi = x$ vanishes for all $x$. Calculating this derivative, we have that $\frac{\partial L(\phi,\chi)}{\partial \phi}$ is
\begin{align*}
     & \equiv \lim_{h \to 0} \frac{L(\phi,\chi) - L(\phi + h,\chi)}{h} \\
    & =  \mathsmaller{\lim_{h \to 0} \mathbb{E}\left[ \frac{(Y - T (\phi+h))^{2} - (Y - T \phi)^{2}}{h\pi(T,X)} \bigg| X = \chi \right]} \\
    & = \mathbb{E}\left[\frac{-2YT + 2\phi}{\pi(T,X)} \bigg| X = \chi \right].
\end{align*}
Then by Euler-Lagrange, we see that $f^{\text{opt}}$ satisfies
\begin{align*}
    0 = & \mathbb{E}\left[\frac{-YT + f^{\text{opt}}(x)}{\pi(T,X)} \bigg| X = x\right] \\
    = & \mathbb{E}_{T}[ \mathbb{E}\left[\frac{-YT + f^{\text{opt}}(x)}{\pi(T,X)} \bigg| T,X = x\right] ] \\
    = & \mathbb{E}_{T}[\frac{1}{\pi(T,x)}(f^{\text{opt}}(x)-\mathbb{E}[YT| T,X=x])] \\
    = & \sum_{t=-1, 1}  (f^{\text{opt}}(x)-\mathbb{E}[YT| T=t,X=x]) \\
    & \times\frac{ P(T=t|X = x)}{\pi(t,x)} \\
    = & 2f^{\text{opt}}(x) -\mathbb{E}[Y^{(1)}| T=1,X=x] \\
    & +\mathbb{E}[Y^{(-1)}| T=-1,X=x] \tag{SUTVA} \\
    = & 2f^{\text{opt}}(x) -\mathbb{E}[Y^{(1)}| X=x] \\
    & + \mathbb{E}[Y^{(-1)}| X=x]. \tag{str. ignor.} \\
    & \implies f^{\text{opt}}(x) = \Delta(x).
\end{align*}
Thus, $f^{\text{opt}}(x)$ is an optimal scoring function and $-\text{sign}(f^{\text{opt}}(x))$ is an optimal ITR.
\end{proof}

In general, there can be many optimal scoring functions for any joint distribution $X,Y,T \sim \mathbb{P}$. For example, for any transformation $h$ that is positive on $\mathbb{R}_{>0}$ and negative on $\mathbb{R}_{<0}$, $\text{sign}(h(\tau(x))$ is optimal for the value function when there is no assumption on the class of rules. This lemma gives a means for identifying one such scoring function. We note that restricting the class of scoring functions to be linear does not mean that we assume the underlying CATE is linear.

Choosing the class of scoring functions/decision rules is usually more a question about policy that clinical practitioners are willing to implement, rather than modeling choice. A linear CATE implies that there is a linear scoring function that matches the unconstrained best value function. But there are also other (nonlinear) scoring functions that can also achieve the best value function. In general, a nonlinear CATE means that a linear scoring function cannot achieve the optimal value function, but the set of best linear decision rules is well defined.

\section{Optimization}
In this section, we describe how to solve optimization problems \eqref{empiricWsqwpen} and \eqref{empiricAsqwpen} when using weighted learning or A-learning as the base learners. We will also discuss how to solve them in a distributed setting but to first fix ideas, we first consider solving these without any data-sharing constraints. The main idea of a block coordinate minimization scheme is to take a Newton step in each coordinate block by minimizing the following objective for each coordinate block $\beta_{j}$:
\begin{align*}
\min_{\beta_{j}} & L(\beta^{cur}) + {U^{cur}}^{\top}(\beta-\beta^{cur})
\\
&
+ (\beta-\beta^{cur})^{\top}\frac{H}{2}(\beta-\beta^{cur}) + \lambda P(\beta),
\end{align*}
Here $\beta^{cur}$ are the current estimates of $\beta$, $U^{cur}$ is the gradient vector at the current estimates, $\frac{\partial L(\beta^{cur})}{\partial \beta}$, and $H$ is the Hessian matrix, $\frac{\partial^{2} L}{\partial \beta \partial \beta^{\top}}$, which happens to be fixed when using weighted learning or A-learning as the base learners. Here $\beta=\beta^{cur}$ outside of the $j$th block. This can be thought of as the second-order Taylor series expansion of the loss around $\beta^{cur}$. Since the blocks outside of the $j$th block are fixed, we can minimize it via the following approximation of the Hessian:

\begin{align}
\min_{\beta_{j}} & L(\beta^{cur}) + {U^{cur}_{j}}^{\top}(\beta_{j}-\beta^{cur}_{j}) \nonumber
\\
&
+ \frac{\gamma_{j}}{2}||\beta_{j}-\beta^{cur}_{j}||^{2} + \lambda P_{j}(\beta_{j}), \label{jtaylorloss}
\end{align}
where $U_{j}^{cur}$ is the gradient with respect to the $j$th coordinate block evaluated at $\beta^{cur}$ and $\gamma_{j}$ is the largest eigenvalue of the $j$th block of $H$. We can compute the gradients $U^{cur}_{Wj}=\frac{\partial L_{W}(\beta^{cur})}{\partial \beta_{j}}$ explicitly for weighted learning as
\begin{align}
\begin{bmatrix}
-\frac{2}{n^{1}} \sum_{i=1}^{n^{1}} w^{1}_{i}X^{1}_{ij}(T^{1}_{i}Y^{1}_{i}-\langle \beta^{1,cur} , X^{1}_{i} \rangle) \\
\vdots \\
-\frac{2}{n^{K}} \sum_{i=1}^{n^{K}} w^{K}_{i}X^{K}_{ij}(T^{K}_{i}Y^{K}_{i}- \langle \beta^{K,cur} , X^{K}_{i} \rangle)
\end{bmatrix} 
\end{align}
where the weights correspond to either the inverse propensity or energy balancing weights discussed in \appendixref{appendix:weighting}. The corresponding gradient, $U^{cur}_{Aj}=\frac{\partial L_{A}(\beta^{cur})}{\partial \beta_{j}}$, for the A-learning method is
\begin{align}
\begin{bmatrix}
-\frac{2}{n^{1}} \sum_{i=1}^{n^{1}} v^{1}_{i}X^{1}_{ij}(Y^{1}_{i}-v^{1}_{i} \langle \beta^{1,cur} , X^{1}_{i} \rangle) \\
\vdots \\
-\frac{2}{n^{K}} \sum_{i=1}^{n^{K}} v^{K}_{i}X^{K}_{ij}(Y^{K}_{i}-v^{K}_{i} \langle \beta^{K,cur} , X^{K}_{i} \rangle)
\end{bmatrix} \label{UA}
\end{align}
where $v_{i}^{k} = (T_{i}^{k}+1)/2 - \pi(1,X_{i}^{k})$ The Hessian approximations that we use are
$\gamma_{Wj} = \max_{k \in [K]} \{ \frac{1}{n^{k}} \sum_{i=1}^{n^{k}} w_{i}^{k}(X^{k}_{ij})^{2} \}$ 
and 
$\gamma_{Aj} = \max_{k \in [K]} \{\frac{1}{n^{k}} \sum_{i=1}^{n^{k}} (v_{i}^{k}X^{k}_{ij})^{2} \} $. Based on a similar argument in \citet{qian2016tweedie}, minimizing problem \eqref{jtaylorloss} yields a decrease in the loss function at each step. By checking the KKT conditions, the solution of problem \eqref{jtaylorloss} is
\begin{align}
\beta^{k,new}_{j} = & \left(\beta^{k,cur}_{j} - \frac{U^{k,cur}_{j}}{\gamma_{j}} \right) \\
& \times \left(1-\frac{\lambda}{|| \varphi_{k}(\gamma_{j}\beta^{cur}_{j} - U^{cur}_{j})||} \right)_{+}, \label{eq:newtonstep}
\end{align}
where $\varphi_{m}(v)$ returns the component-wise positive or negative part of $v$, according to the sign of the $m$th element (if it is positive or negative).

Now we consider the perspective of computing this step in a distributed setting. In our setup, we assume that communication occurs in rounds: within each round, sites can exchange messages with the centralized coordinating center (spoke and wheel topology) and between each round, sites compute based on the data at the site and information received from previous rounds. For data privacy reasons, we cannot share data or models between sites.

Clearly, the usual strategy (e.g. federated learning) of transferring gradients $U_{j}$ from sites to servers allows us to compute this step at a central server. However, in the case where sharing of parameters and gradients is sensitive \citep{wang2020attack, bagdasaryan2020backdoor, geiping2020inverting}, we note that to compute \eqref{eq:newtonstep} locally, only the shrinkage term depends on information from other sites. Thus, it is sufficient to compute this step by only transferring $db^{k}_{j} \equiv \gamma_{j}\beta_{j}^{k} - u_{j}^{k}$ to the central server, which in turn broadcasts the step sizes 
\begin{align}
    \left(1-\frac{\lambda}{|| (\gamma_{j}\beta^{cur}_{j} - U^{cur}_{j})_{+}||} \right)_{+}, \\
    \left(1-\frac{\lambda}{|| (\gamma_{j}\beta^{cur}_{j} - U^{cur}_{j})_{-}||} \right)_{+} \label{stepsize}
\end{align}
back to each site, depending on the sign of the $k$th bit received. This motivates the algorithm presented in \algorithmref{proc:optimization}. This algorithm achieves the optimal communication lower bound of $O(pK)$ bits per iteration as in the homogeneous populations, homogeneous concept case \citep{garg2014communication}.

\begin{algorithm2e}
\label{proc:optimization}
\caption{Optimization for individual-level meta-analysis}
		\KwIn{Data from each site $\{(Y_{i},X_{i},T_{i})\}_{i \in [n_{k}]}$ for $k \in [r]$}
		\KwOut{ITR coefficients from each site $\beta^{1}, \ldots, \beta^{K}$}
		\For{Site $k$ in $[K]$}{
		    Compute site-specific weights $\{w_{i}\}^{k}$ and augmentation $\hat{a}^{k}$ \;
		    Initialize $\beta^{k,0}$ to solution to local subproblem $L^{k}_{W}$ \;
		    Compute Hessian eigenvalues $\gamma^{k}_{j}$ for all $j \in [p]$ \;
		    Broadcast $p$ eigenvalues to central server \;
		}
		Approximate Hessian for all $j$ by setting $\gamma_{j} = \max_{k}\gamma^{k}_{j}$ \;
		Broadcast $\gamma_{j}$ for $j \in [p]$ to all sites $k \in [K]$ \; 
		\For{$t = 0, 1, \ldots, T-1$}{
            \For{Feature $j$ in $[p]$}{
                \For{Site $k$ in $[K]$}{
                    Compute the $k$th element of $U_{Wj}$, $u^{k}_{Wj} = -\frac{2}{n^{k}} \sum_{i=1}^{n^{k}} w^{k}_{i}X^{k}_{ij}(T^{k}_{i}Y^{k}_{i}-\langle \beta^{k,t} , X^{k}_{i} \rangle)$ \; 
    		        Transmit scalar $db^{k}_{j} = \gamma_{j}\beta^{k}_{j} - u^{k}_{Wj}$ to central server \; 
                }
		            At central server, stack $dB_{j}  = \begin{bmatrix}
		                db^{1}_{j}  & \ldots & db^{K}_{j} 
		            \end{bmatrix}^{\top}$ \; 
		            Broadcast step size \eqref{stepsize} back to each site $k$, depending on if $db^{k}_{j}$ was positive or negative. \; 
		            \For{Site $k$ in $[K]$}{
                        Compute $\beta^{k,t+1}_{j}$ using equation \eqref{eq:newtonstep} \;
                    }
            }
    		\tcc{finished one cycle of coordinates}
        }
\end{algorithm2e}

In our implementation, we also use a strong rules approach to discarding predictors \citep{tibshirani2012strong} for efficient optimization, so for each iteration, only a small `active' fraction of the $p$ predictors need to be cycled through. The worst case complexity of using strong rules does not yield any improvements, but in practice, we find that the optimization with strong rules is quite fast. We give the details of the strong rules filtering in the next section.

Compared to a federated learning approach \citep{mcmahan2017communication}, each site only needs to transfer a scalar during each communication round, instead of transferring gradients. For added security, a drop-in differential privacy method could be used when transmitting that scalar \citep{abadi2016deep}.

\section{Strong rules for cooperative LASSO}
When computing model coefficients along a solution path of a hyperparameter $\lambda_{\ell}$ for $\ell \in L$, the strong rule strategy \citep{tibshirani2012strong} is to use a heuristic to discard a large amount of predictors at each step. This is not exact, so sometimes a mistake is made and a nonzero predictor is discarded. However, this is so rare that discarding according to the strong rule first and then checking the KKT conditions for mistakes later results in a beneficial computational trade-off. In this section, we use the notation $X = \begin{bmatrix}
    X^{1} &  &0\\
    & \ddots  & \\
    0 & & X^{k}
\end{bmatrix}$, $Y = \begin{bmatrix}
    Y^{1\top} & \ldots & Y^{K\top}
\end{bmatrix}^{\top}$, $\beta = \begin{bmatrix}
    \beta^{1\top} & \ldots & \beta^{K\top}
\end{bmatrix}^{\top}$ to match the strong rules literature. We let $\mathcal{G}_{j}$ be the group of $j$th coefficient indices from each site. Following a similar derivation as \citet{tibshirani2012strong}, we start from the KKT conditions of the cooperative LASSO,
\begin{align*}
    \langle X,Y-X\beta \rangle = \lambda \boldsymbol{\theta}
\end{align*}
where $\theta$ is a subgradient of the coop norm. We can write out the $j$th index explicitly as $\langle X_{j}, Y-X \beta \rangle = \lambda \theta_{j}$. By Theorem 1 of \citet{chiquet2012sparsity}, we have

\begin{align*}
     \langle x_{k}, y- & X\beta \rangle =  \frac{\lambda \beta_{k}}{ || \varphi_{k}(\beta_{\mathcal{G}_{j}}) || } \\
    & \text{ for $k \in \mathcal{G}_{j}$ and $\beta^{k}_{j}$ is nonzero} \\
     || \varphi_{k} ( \langle & X_{\cdot \mathcal{G}_{j}} , Y - X\beta \rangle ) || \leq \lambda \\
    & \text{ for $k \in \mathcal{G}_{j}$ and $\beta^{k}_{j}$ is zero}
\end{align*}

We let $c_{kj}(\lambda) = \varphi_{k}( \langle X_{\cdot \mathcal{G}_{jgroup}}, Y-X\hat{\beta}(\lambda) \rangle )$ be a function of $\lambda$. The key approximation for strong rules is that we assume that $c_{kj}$ is nonexpansive with respect to the 2-norm:
\begin{align*}
    ||c_{kj}(\lambda) - c_{kj}(\tilde{\lambda}) || \leq |\lambda - \tilde{\lambda} ||.
\end{align*}
If we choose our strong rule to be
\begin{align}
    ||  \varphi_{k}( \langle X_{\cdot \mathcal{G}_{jgroup}} &, Y-X\hat{\beta}(\lambda_{\ell-1}) \rangle ) || \nonumber \\
    < & 2 \lambda_{\ell} - \lambda_{\ell-1}. \label{strongrule}
\end{align}
Then we get
\begin{align*}
    || c_{kj}(\lambda_{\ell}) || \leq & || c_{kj}(\lambda_{\ell}) - c_{kj}(\lambda_{\ell-1}) || \\
    & + || c_{kj}(\lambda_{\ell-1}) || \\
    < & (\lambda_{\ell-1} - \lambda_{\ell}) + (2 \lambda_{\ell} - \lambda_{\ell-1}) \\
    = & \lambda_{\ell},
\end{align*}
implying that $\hat{\beta}^{k}_{j}(\lambda_{\ell})$ vanishes. In the strict inequality above, we used the nonexpansive assumption for the first term, strong rule for the second term.

\section{Cross-fitting}
\label{appendix:crossfit}
As we have seen, estimation of the ITR depends on both an estimated propensity score and augmentation function. These need to be estimated from the data, but reusing data for estimating these and for fitting the scoring function can lead to overfitting \citep{chernozhukov2017double}. In order to overcome this limitation, we use a cross-fitting strategy \citep{zheng2011cross} that decouples the estimation of the ITR from the estimation of the propensity score and augmentation term. Within a given site, samples are split into folds so that the augmentation and propensity functions evaluated on a given sample point were trained on out-of-fold data.

\section{Energy balancing weights}
\label{appendix:weighting}
The core reason that the learning problem needs to be cast as a weighted regression problem is to achieve a valid causal estimate by achieving covariate balance of confounding variables. There are two major approaches to achieving covariate balance. The first is to fit a propensity score and then invert it, as discussed previously. An alternative is to use a weighting method that achieves covariate balance by directly optimizing for balance \citep{ben2021balancing}. We can instead minimize $\frac{1}{n} \sum_{i=1}^{n} w(T_{i},X_{i})(Y_{i}- \hat{a}(X_{i}) - T_{i}f(X_{i}))^{2}$ as the weighted learning loss function, where $w(t,x)$ is a weighting function that achieves approximate covariate balance: 
\begin{equation}
    \mathbb{E}\left[ w(T,X)g(T,X) \right]= \mathbb{E}\left[ \frac{g(T,X)}{\pi(T,X)} \right]
\end{equation}
for any integrable function $g$. In the multi-site setting, achieving balance \textit{within} sites is sufficient for a valid approximation of the multi-site loss function \eqref{multistudyloss}. Many approaches attempt to balance covariate distributions by balancing a finite number of moments of the covariates, but this trades careful modeling with careful hyperparameter tuning. The energy balancing weight approach \citep{huling2020energy} overcomes this by using a weighted energy distance metric on the distributions to avoid both model specification and hyperparameter tuning, achieving a causal estimate by direct balancing of the distributions of confounders across treatment groups. Fundamentally, the choice of balancing method is contextual, and our implementation provides options for parametric and nonparametric propensity-score fitting and energy balancing weighting.

\section{Adaptive information criteria}
\label{appendix:adaptive}
The concordance function is a population-level quantity defined for single sites as
\begin{align*}
    C(f) = \mathbb{E} [ & \left( \Delta(X) - \Delta(X') \right) \\
    & \times I(f(X) < f(X')) ],
\end{align*}
where $X_{i}$ and $X_{i'}$ are two independent copies of the features. The interpretation of concordance is that it measures the compatibility between a scoring function and the actual treatment contrast.

In turn, the (single site) concordance information criterion is defined as $\text{CIC}_{\kappa_{n} }(\beta) = n \hat{C}(\beta) - \log (n)||\beta||_{0}$, where $\hat{C}(\beta)$ is the augmented inverse probability weighting (AIPW) estimator \citep{bang2005doubly} of the concordance
\begin{align}
	& \sum_{i \neq j} \bigg\{ \big[  w_{i}(Y_{i} - \hat{a}(X_{i})) -  w_{j}(Y_{j} - \hat{a}(X_{j})) \big] \nonumber \\
	& \times \mathbb{I}(\langle \beta, X_{i} \rangle > \langle \beta, X_{j} \rangle ) \bigg\}\frac{1}{n(n-1)} , \label{multiconcordance}
\end{align}
and the 0-norm $||\cdot ||_{0}$ is the cardinality of nonzero elements \citep{shi2021concordance}. The weights $w_{i}$ can be any covariate balancing weights as discussed in \appendixref{appendix:weighting}. The form of this information criteria resembles the Bayesian Information Criterion (BIC), with the log-likelihood replaced by estimates of the concordance or value function respectively.

One obstacle in directly using CIC (VIC) is dependence on the choice of outcome standardization. In ordinary practice, researchers often make a choice to rescale or transform variables, for reasons of interpretability or computational stability. From the forms above, we note that multiplying $Y$ by a constant leads to a corresponding multiplication in the fitted augmentation term by the constant in the concordance term in CIC (VIC) while leaving the complexity penalization unchanged. While this is not an issue for the asymptotic theory, this ambiguity in how to scale relative terms in CIC (VIC) leads to ambiguity in model selection in practice.

In order to resolve this ambiguity and make our method robust to the choice of outcome scaling, we develop a data-adaptive form of CIC (VIC). The intuition behind the approach is that CIC (VIC) defines a single curve in $\lambda$. With an auxiliary scaling parameter, we can then define a family of curves, indexed by $\gamma$:
\begin{equation}
    CIC(\lambda, \gamma) = n\hat{C}(\hat{\beta}_{\lambda}) - \gamma\log(n)||\hat{\beta}_{\lambda}||_{0} \label{aCIC}.
\end{equation}
A good choice of $\gamma$ for scaling results in a curve with high discrimination between the best model and other models. In the case where curves are weakly differentiable, the total variation of a curve indicates how `sharp' its derivatives are. Motivated by this, we propose Procedure \ref{proc:adaptiveCIC}.

\begin{algorithm2e}
\caption{Adaptive CIC (VIC)}
\label{proc:adaptiveCIC}
\KwIn{Sequence of models $\hat{\beta}(\lambda)$ for $\lambda \in \Lambda$}
\KwOut{Value of $\lambda^{\text{opt}}$ and model $\hat{\beta}(\lambda^{\text{opt}})$}
\For{Sequence $\gamma_{min} \leq \gamma \leq \gamma_{max}$}{
    Compute $CIC(\lambda, \gamma)$ (or $VIC(\lambda, \gamma)$) according to eq. \eqref{aCIC} \;
    Set $\tilde{CIC}(\lambda, \gamma)$ be the normalized curve \;
    Compute $TV(\tilde{CIC}(\lambda, \gamma))$ \;
}
Pick the curve with the greatest total variation norm \;
Pick the value of $\lambda$ that maximizes this curve \;
\end{algorithm2e}

The interpretation is that the CIC curve with the largest total variation norm, after renormalizing, corresponds to the curve with the best model discrimination. We implement this numerically via first-order finite differences.

\section{Simulation data generation}
\label{appendix:simulation}
For the simulations only, we assume larger $Y$ is better. Since the data is generated, we can directly compare the accuracy of treatment rules estimated to the true sign of the treatment effect, as well as compute a relative value function, which is a fraction of the maximum potential value captured by the treatment rule. Within a single replication, the data within a site is generated by a model and noise according to
\begin{align*}
    Y^{k}_{i} = & \mu^{k}(X^{k}_{i}) + T^{k}_{i}\Delta^{k}(X^{k}_{i}) +  \varepsilon^{k}_{i} \\
     = & \overbrace{ \underbrace{ \langle \alpha^{k}, X^{k}_{i}\rangle }_{\text{linear terms}} + \underbrace{ \langle \gamma^{k}, Z_{i}^{ k } \rangle }_{\text{quadratic terms}} }^{\text{main effect}} \\
    & +  T^{k}_{i} \overbrace{( \underbrace{ \langle \beta^{k}, X^{k}_{i} \rangle }_{\text{linear terms}} + \underbrace{ \langle \zeta^{k}, Z_{i}^{ k } \rangle ) }_{\text{quadratic terms}} }^{\text{treatment effect}} +\varepsilon^{k}_{i}.
\end{align*}
where $X_{i}^{k}$ is standard multivariate normal, $Z_{i}^{ k } = \text{vec}(X_{i}^{ k }\otimes X_{i}^{ k })$ is the vector of second order terms of covariates, $\begin{bmatrix}
    \alpha^{ k }
\end{bmatrix}$ and  $\begin{bmatrix}
    \beta^{ k }
\end{bmatrix} \in \mathbb{R}^{(p+1) \times 1}$ are vectors of parameters (including intercepts) for the linear main and interaction effects, respectively, and $\gamma^{k}, \zeta^{k} \in \mathbb{R}^{(p+1)^{2} \times 1}$ are vectors of parameters for the nonlinear main and interaction effects, respectively. The sparsity of coefficients is set so 15\% of the $p$ elements of $\alpha$ and $\beta$ are nonzero in each replication. The sparsity of coefficients for the nonlinear terms is set so 8\% of the $(p+1)^{2}$ elements are nonzero. The support of these coefficients is drawn uniformly at random for each replication. Each nonzero linear parameter and intercept is the sum of a shared `base' component and a site-specific `perturbation' component, multiplied by a sign:
\begin{align*}
    \alpha_{j}^{\{ k \}} & = s_{\alpha, j}(\alpha_{j,\text{base}} + \alpha^{\{ k \}}_{j,\text{perturb}}), j \in [p]\cup \{ 0\} \\
    \beta_{j}^{\{ k \}} & = s_{\beta, j}(\beta_{j,\text{base}} + \beta^{\{ k \}}_{j,\text{perturb}}), j \in [p]\cup \{ 0\}.
\end{align*}
Each sign is a Rademacher random variable and ensures sign-coherency of the linear coefficients between studies. The nonzero base and perturbation coefficients are all drawn from a log-normal($\mu=0,\sigma^{2}=1$) distribution in the first scenario, hence a balance between homogeneity and heterogeneity. For the nonlinear coefficients, there are no site-specific perturbation terms, and the nonzero components are drawn from lognormal($0,1$). The error term is generated from a standard normal independently of all other parameters. Each of the 500 replications for each sub-scenario is performed under a fresh set of parameters.

In the second scenario, the site-specific perturbation coefficients for the linear interaction effects are drawn from lognormal($5,1$) while the base coefficients are still drawn from lognormal($0,1$) so the site-specific component is larger than the shared component. In the third scenario, this is flipped with the base coefficients for the linear interaction effects drawn from lognormal($5,1$) while the site-specific perturbation is drawn from lognormal($0,1$) so the between-site heterogeneity is larger. In all sub-scenarios, the test set is 10,000 points generated from the above model, including both potential outcomes $T=1, T=-1$.

\begin{figure*}
    \centering
    {\includegraphics[width=0.95\linewidth]{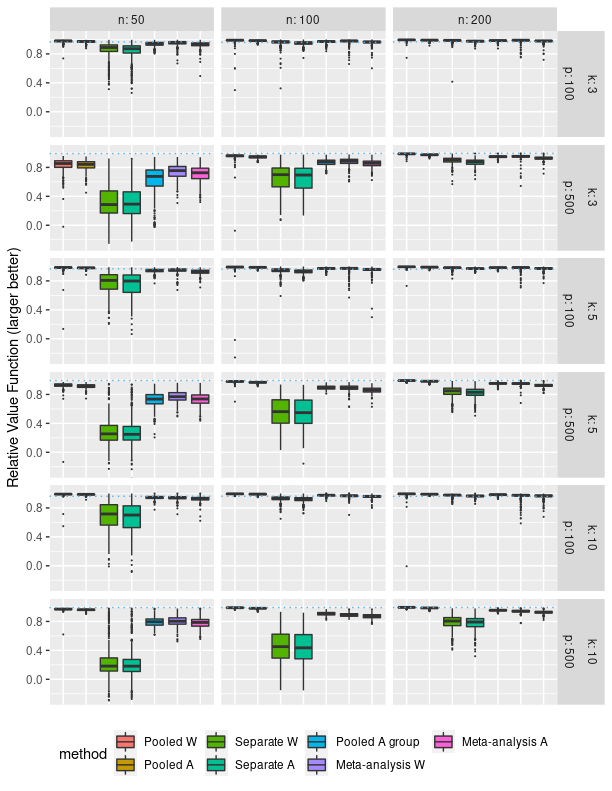}}
    \caption{Relative value function under low site heterogeneity.}
    \label{fig:50val}%
\end{figure*}

\begin{figure*}
    \centering
    {\includegraphics[width=0.95\linewidth]{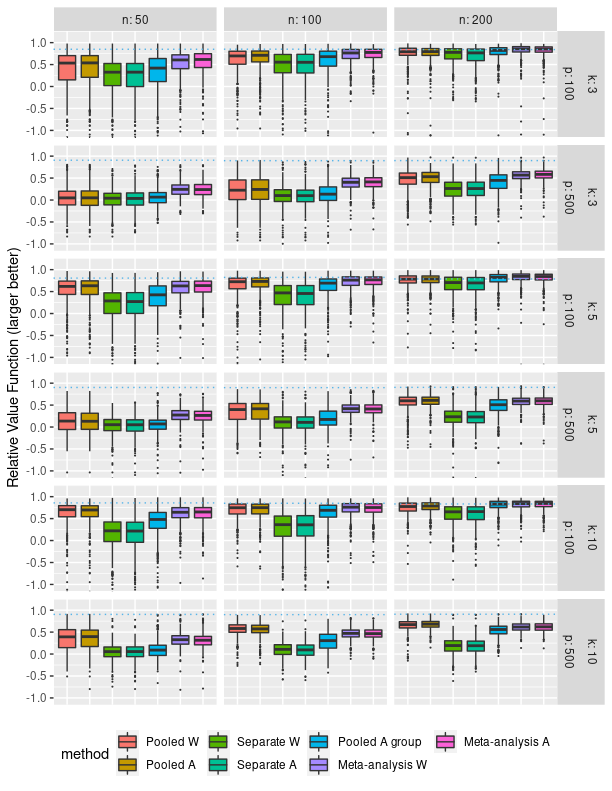}}
    \caption{Relative value function under moderate site heterogeneity.}
    \label{fig:00val}%
\end{figure*}

\begin{figure*}
    \centering
    {\includegraphics[width=0.95\linewidth]{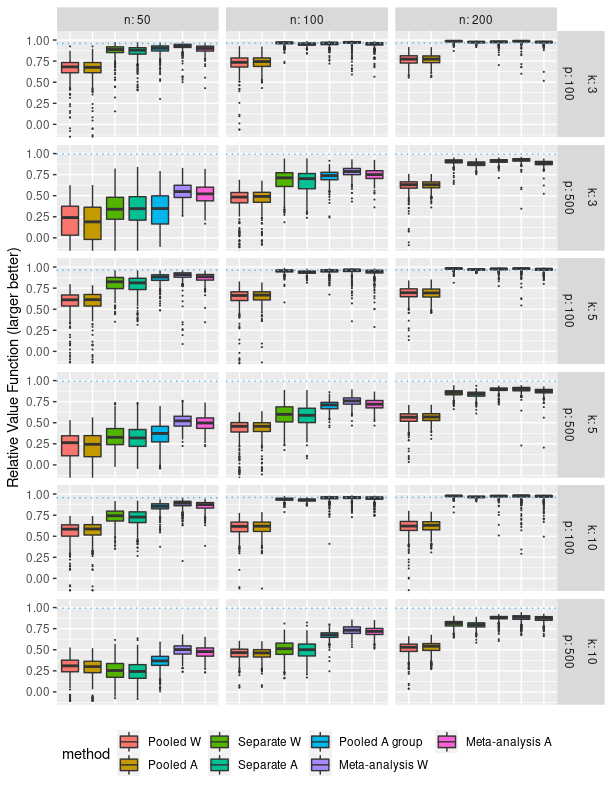}}
    \caption{Relative value function under high site heterogeneity.}
    \label{fig:05val}%
\end{figure*}

\begin{figure*}
    \centering
    {\includegraphics[width=0.95\linewidth]{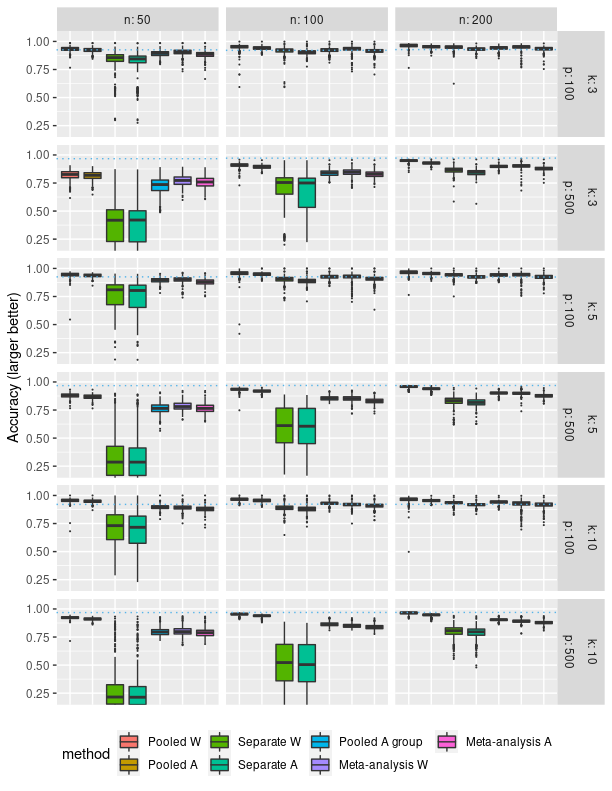}}
    \caption{Treatment rule accuracy compared to true potential outcomes under low site heterogeneity.}
    \label{fig:50acc}%
\end{figure*}

\begin{figure*}
    \centering
    {\includegraphics[width=0.95\linewidth]{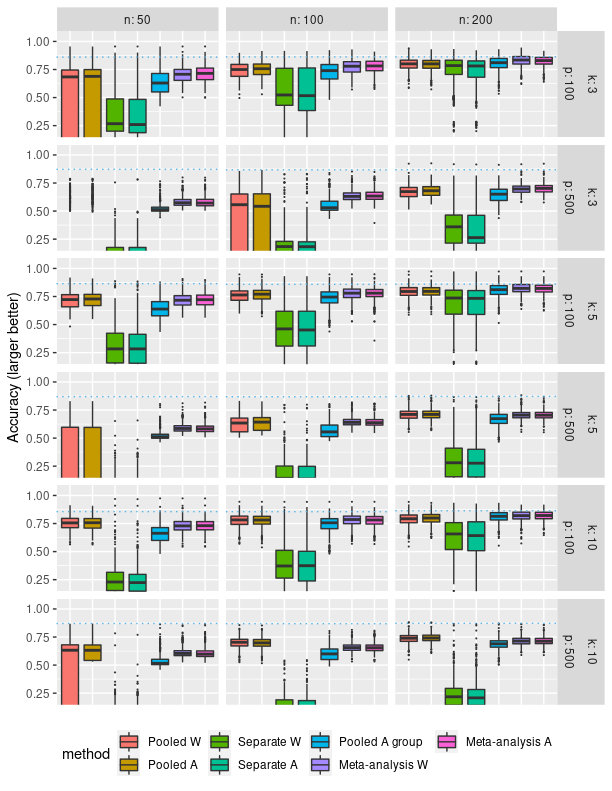}}
    \caption{Treatment rule accuracy compared to true potential outcomes under moderate site heterogeneity.}
    \label{fig:00acc}%
\end{figure*}

\begin{figure*}
    \centering
    {\includegraphics[width=0.95\linewidth]{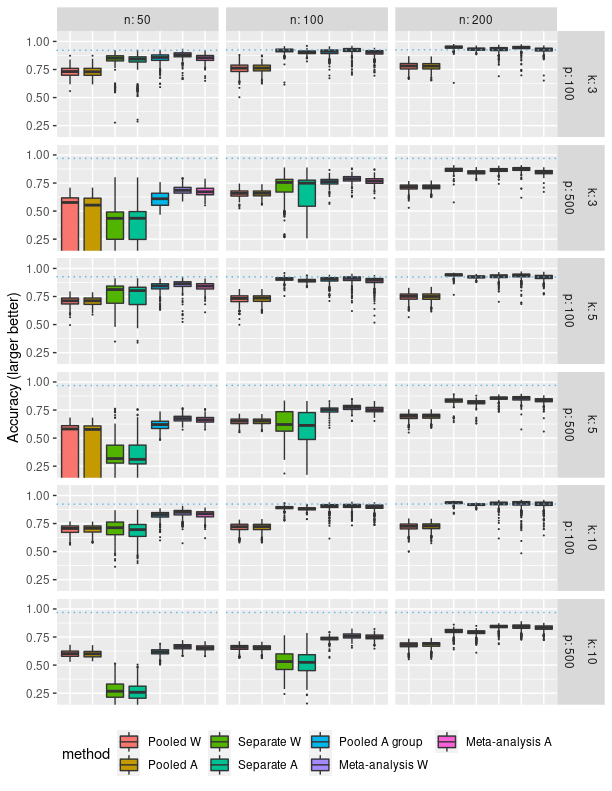}}
    \caption{Treatment rule accuracy compared to true potential outcomes under high site heterogeneity.}
    \label{fig:05acc}%
\end{figure*}

\section{Sensitivity to unmeasured confounders}
While we take ignorability as a starting point, sensitivity of the model to unmeasured confounders is an important consideration before real clinical decisions are made. Sensitivity analysis is an important research topic in causal inference. However, most sensitivity analysis focus on the average treatment effect, the ones focus on the ITR are very sparse except \citet{kallus2018confounding} and \citet{zhang2021selecting}. Alternatively, there are some recent attempts to relax this assumption based on instrumental variable framework including \citet{cui2021semiparametric, qiu2021optimal, pu2020estimating}. To understand the sensitivity of meta-analysis to unmeasured confounders, we conducted additional simulations, repeated for 500 replications, where 1/3 of the variables with true effects are missing.

\begin{table}[h!]
    \centering
     \caption{First quartile, mean, and third quartile of relative value function compared to the maximum potential outcome over 500 replications for many missing confounders. MA=metaanalysis, BLDR=best linear decision rule.}
    \begin{tabular}{c|c|c|c}
    \hline
        & 1Q & mean & 3Q \\
        \hline
         pooled W & 0.3814 & 0.4554 & 0.6290 \\
         pooled A & 0.3682 & 0.4479 & 0.6263  \\
         separate W & 0.4204 & 0.5735 & 0.7662 \\
         separate A & 0.4194 & 0.5754 & 0.7673 \\
         group A & 0.4364 & 0.5811 & 0.7813 \\
         MA W & 0.5586 & 0.6688 & 0.8176 \\
         MA A & 0.5802 & 0.6824 & 0.8219 \\
         BLDR & 0.6878 & 0.7614 & 0.8622 \\
         \hline
    \end{tabular}
\end{table}

All methods are sensitive to unmeasured confounders, but it seems like metaanalysis is the most robust to unmeasured confounding.

\section{Performance under Covariate Shift}
Following the same data generation mechanism for $n=50$, $p=100$, and $K=3$ we performed additional simulations where every covariate in each site are randomly shifted up to one standard deviation compared to the original data generation process. For 500 replications, we present the 1st quartile, mean, and 3rd quartile of the relative value captured.

\begin{table}[h!]
    \centering
    \caption{First quartile, mean, and third quartile of relative value function compared to the maximum potential outcome over 500 replications with large covariate drift between sites. MA=metaanalysis, BLDR=best linear decision rule.}
    \begin{tabular}{c|c|c|c}
    \hline
        & 1Q & mean & 3Q \\
        \hline
         pooled W & 0.6352 & 0.6786 & 0.7629 \\
         pooled A & 0.6352 & 0.6845 & 0.7616  \\
         separate W & 0.8830 & 0.9028 & 0.9355 \\
         separate A & 0.8631 & 0.8866 & 0.9264 \\
         group A & 0.8720 & 0.8921 & 0.9327 \\
         MA W & 0.9223 & 0.9360 & 0.9550 \\
         MA A & 0.8982 & 0.9155 & 0.9396 \\
         BLDR & 0.9620 & 0.9607 & 0.9940 \\
    \hline
    \end{tabular}
\end{table}

\section{Imputation for Censored Discharge Times}\label{supp:censimpute}
Let the true discharge time (in absence of censoring) be denoted $\tilde{Y}$ and the censoring time be denoted $C$. In this case, the censoring time is the time of death. We use $\tau$ to denote the maximum observable time. Let $\delta=\mathbb{I}(\tilde{Y} < C)$ be the indicator that the time we observe is either a censoring time or survival time. The observed time is $Y=\min(\tilde{Y},C)$, either censored time or survival time. Our setup is that we observe $(Y, \delta, X, T)$ where $X$ are the pre-treatment covariates and $T$ is the treatment.

To address the censoring bias, we impute a discharge time to the patients who died. The imputation approach follows the approach taken in \citet{leete2019balanced}. For missing discharge times, we impute it with an estimate of the conditional average discharge time, $\hat{\mathbb{E}}[\tilde{Y} | X, T, Y, \delta = 0]$,
\begin{align*}
    \frac{\tau \hat{S}(\tau | X,T) + \int_{Y}^{\tau} t \hat{dF}(t|X,T)}{\hat{S}(Y|X,T)},
\end{align*}
where $\hat{S}$ is an estimate of the survival function and $\hat{dF}$ is an estimate of the density function of $\tilde{Y}$. We estimate $\hat{S}$ with random survival forests and use the relation $\hat{dF}(t|X,T) = \mathbb{P}(t \leq T < t + dt | X,T)$ to integrate.

\end{document}